\documentclass[conference]{IEEEtran}  
\IEEEoverridecommandlockouts
\usepackage{cite}
\usepackage{amsmath,amsfonts,amsthm,bm} % Math packages
\usepackage{caption}
\usepackage{subcaption} %To insert subfigures 
\usepackage{verbatim}  %To comment out large sections of text with \begin{comment} and \end{comment}
\usepackage{colortbl} %For using colors in header of tables
\usepackage{algorithm}
\usepackage{algpseudocode}
\usepackage{graphicx}
\usepackage{textcomp}
\usepackage{xcolor} 
\usepackage{array,multirow} %For changing borders in tables

\usepackage{tabularx,booktabs}
\newcolumntype{C}{>{\centering\arraybackslash}X} % centered version of "X" type
\definecolor{maroon}{cmyk}{0,0.87,0.68,0.32} %For headers in table instead of gray (it has different intensities, for example: !10 has less intensity)
\usepackage{flushend} %Align last page (same length both columns)
\usepackage{url} 
\usepackage{authblk} %To add authors from different departments and universities

\usepackage{lipsum}  %Generate dummy text
\usepackage{amssymb} % To use symbol greater or equal approximately: \gtrapprox

\usepackage{mathtools}

\def\BibTeX{{\rm B\kern-.05em{\sc i\kern-.025em b}\kern-.08em
    T\kern-.1667em\lower.7ex\hbox{E}\kern-.125emX}}
\begin{document}

\title{Study of the gOMP Algorithm for Recovery of Compressed Sensed Hyperspectral Images}

\author[]{Jon Alvarez Justo}
\author[]{Milica Orlandić}

\affil[]{Department of Electronic Systems, Norwegian University of Science and Technology (NTNU)}

\renewcommand\Authands{, and }

\maketitle

\begin{abstract} 

Hyperspectral Imaging (HSI) is used in a wide range of applications such as remote sensing, yet the transmission of the HS images by communication data links becomes challenging due to the large number of spectral bands that the HS images contain together with the limited data bandwidth available in real applications. Compressive Sensing reduces the images by randomly subsampling the spectral bands of each spatial pixel and then it performs the image reconstruction of all the bands using recovery algorithms which impose sparsity in a certain transform domain. Since the image pixels are not strictly sparse, this work studies a data sparsification pre-processing stage prior to compression to ensure the sparsity of the pixels. The sparsified images are compressed $2.5\times$ and then recovered using the Generalized Orthogonal Matching Pursuit algorithm (gOMP) characterized by high accuracy, low computational requirements and fast convergence. The experiments are performed in five conventional hyperspectral images where the effect of different sparsification levels in the quality of the uncompressed as well as the recovered images is studied. It is concluded that the gOMP algorithm reconstructs the hyperspectral images with higher accuracy as well as faster convergence when the pixels are highly sparsified and hence at the expense of reducing the quality of the recovered images with respect to the original images. 

\end{abstract}


\section{Introduction}
\footnotetext{IEEE-COPYRIGHTED MATERIAL - © 2022 IEEE. Personal use of this material is permitted.
Permission from IEEE must be obtained for all other uses, in
any current or future media, including reprinting/republishing
this material for advertising or promotional purposes, creating
new collective works, for resale or redistribution to servers or
lists, or reuse of any copyrighted component of this work in
other works.}

Hyperspectral Imaging (HSI) collects and processes light across hundreds of narrow spectral bands in the electromagnetic spectrum providing valuable information for aquaculture, agriculture, food analysis, oil and remote sensing applications for classifying different types of targets~\cite{HSIapplications}. However, there are challenges for space satellites and airborne platforms such as unmanned aerial systems with hyperspectral equipment on-board to transmit the vast amount of HS data. As an example, the Small Satellite in the mision \textit{Hyperspectral Satellite for Ocean Observation} (HYPSO-1) which monitors ocean phenomena faces some limitations such as windowed transmission times and bounded bandwidth~\cite{HYPSO1paper,SmallSatLab}. The uncompressed HS images captured by HYPSO-1 are approximately $80\;\mathrm{MB}$ to be transmitted by a communication channel of $1\;\mathrm{Mbps}$ bandwidth and within $10\;\mathrm{minutes}$ average transmission time in one orbit pass. As a result, a complete HS image cannot be transmitted in one satellite pass and hence more than one orbit is required to complete the image transmission. Different techniques to compress the HS images proposed in the current state of the art are listed next~\cite{review_compression_HSI}. Transform based methods such as \textit{Weighted Principal Component Analysis} (WPCA)~\cite{WPCA_paper}, compression algorithms from the \textit{Consultative Committee for Space Data Systems} (CCSDS) such as the lossless CCSDS-123 Issue 1~\cite{ccsds123_lossless1,ccsds123_lossless2} and the lossy CCSDS-123 Issue 2~\cite{ccsds123_lossy}, learning based techniques using \textit{Convolutional Neural Network-Lempel-Ziv Markov chain} algorithm (CNN-LZMA) where the compressed HS images are coded by LZMA and reconstructed by a CNN trained through residual learning, \textit{Vector Quantization Principal Component Analysis} (VQPCA)~\cite{VQPCA_paper}, and \textit{Compressive Sensing} (CS)~\cite{CS_paper}. In CS applied to imaging, the compression stage consists of subsampling random image data samples and then decompressing the images using algorithms which perform the recovery in a certain sparse basis where most of the image samples are zeros except from a few significant samples. The preceding paper to this work~\cite{1st_paper_cs_comparison} compares the accuracy and performance of convex and greedy algorithms for recovering subsampled HS images. The paper shows that the convex algorithms \textit{Fast Iterative Shrinkage-Thresholding Algorithm} (FISTA)~\cite{paper_FISTA_cited_from_Ion}/ \textit{Alternating Direction Method of Multipliers} (ADMM)~\cite{paper_ADMM_cited_from_Ion} in addition to the greedy algorithms \textit{Backtracking Iterative Hard Thresholding} (BIHT)~\cite{bib:original_paper_BIHT}/ \textit{Compressive Sampling Matching Pursuit} (CoSaMP)~\cite{bib:original_paper_CoSaMP}/ \textit{Generalized Orthogonal Matching Pursuit} (gOMP)~\cite{bib:original_paper_GOMP} are able to recover the HS images. However, it concludes that the gOMP algorithm is the most efficient in terms of the accuracy measured through the \textit{peak signal-to-noise ratio} ($\mathit{PSNR}$) metric and also regarding the performance given by the highest convergence ratio in addition to the lowest computational requirements. This paper continues the work on the gOMP algorithm performing a study on the sparsification stage prior to compression and analysing how the image quality of the uncompressed and the recovered images is affected for different sparsification levels. The \textit{structural similarity index} ($\mathit{SSI}$) is used as an additional metric to the $\mathit{PSNR}$ for measuring the recovery accuracy.
The remainder of the paper is organized as follows. Section~\ref{Section: background} approaches how the CS technique is applied to the HS images, it presents the gOMP algorithm as well as the different accuracy and performance metrics. Section~\ref{Section: results} presents a sparsity study of different HS pixels in addition to analysing the recovery results of the complete HS images. Finally, Section~\ref{Section: conclusions} concludes and suggests future work. 
 %that the quality metrics such as the \textit{mean squared error} ($\mathit{MSE}$) and the $\mathit{PSNR}$ are simple to calculate while holding convenient optimization properties~\cite{SSIM_orig_paper}, yet showing low correlation with the perceived visual quality of images. Therefore, the literature presents the \textit{Structural Similarity Index} ($\mathit{SSI}$) which consists in a perceptual metric exploiting visible structures in the images, and it has been originally used in JPEG and JEPG2000~\cite{SSIM_orig_paper} but also for HS images \textcolor{blue}{[ssi hsi]} \textcolor{blue}{no percept.}.

% Hypso-1 numbers (from Roger): average pass is around 10 mins, cubes are 70-80 MB, bandwidth is 1 Mbps, and he mentions that in one pass we usually get 60-80\% of the cube. From Dennis: 150 MB uncompressed (compression is lossless CCSDS-123, implemented in either SW or HW)

\section{Background}
\label{Section: background}

\subsection{Compressive Sensing in Hyperspectral Imaging}

A HS spatial pixel $\mathbf{f} \in \mathbb{R}^{N\times 1}$ has $N$ spectral samples and it can be represented with the transform equation $\mathbf{f}=\mathbf{\Psi}\mathbf{x}$ where the transform matrix $\mathbf{\mathbf{\Psi}} \in \mathbb{C}^{N\times N}$ is the \textit{Discrete Fourier Transform} (DFT) matrix and hence the pixel $\mathbf{f}$ is in the DFT acquisition domain while the vector $\mathbf{x} \in \mathbb{C}^{N\times 1}$ is a representation of the pixel $\mathbf{f}$ in the \textit{Inverse Discrete Fourier Transform} (IDFT) domain. The image compression is performed computing the projections of each spatial pixel $\mathbf{f}$ over a different randomized measurement matrix $\mathbf{\Phi} \in \mathbb{R}^{M\times N}$, where $M$ is the number of subsampled spectral bands and the resulting compressed spectral samples are given in the measurement vector $\mathbf{y} \in \mathbb{R}^{M\times 1}$ which can be expressed as follows: 

\begin{equation}
    \mathbf{y}=\mathbf{\Phi}\mathbf{f}=\mathbf{\Phi}\mathbf{\Psi}\mathbf{x}=\mathbf{A}\mathbf{x}+\lambda \| \mathbf{x}\|_1,
\label{eq:main_CS_eq}
\end{equation}

\noindent where the matrix $\mathbf{A} \in \mathbb{C}^{M\times N}$ denotes the \textit{dictionary} whose columns are known as \textit{atoms} and $~\lambda~\in~\mathbb{R}^{1\times 1}~$ is a regularization parameter which imposes sparsity to the vector $\mathbf{x}$ to enforce a unique solution to the underdetermined system of linear equations given in (\ref{eq:main_CS_eq}). The $\kappa$-sparsity of the vector $\mathbf{x}$ is defined as the number of samples different from zero, i.e., $ \kappa\;=\;{\parallel \mathbf{x} \parallel}_0$. The sparsity ratio, denoted as $\mathit{SR}$, measures the degree of sparsity of the pixel $\mathbf{x}$ by calculating the ratio between the number of samples which are zero and the total number of data samples. To ensure that the pixel $\mathbf{x}$ is strictly sparse, a sparsification stage prior to the compression is used where the samples in $\mathbf{x}$ above a certain threshold are maintained while the remaining samples are rounded down to zero when the next condition is satisfied: 
\begin{equation}
    \mid \mathbf{x} \mid\;<\; \frac{T}{100}\cdot \;\max(\mid\mathbf{x}\mid),
\label{eq:sp_condition}
\end{equation}

\noindent where the parameter $T$, given in $\%$, denotes the sparsification threshold and it is adjusted experimentally to achieve the desired sparsity level $\kappa$. For higher $T$ values, more samples in the vector $\mathbf{x}$ are rounded down to 0, that is, smaller $\kappa$ is obtained and hence increasing the $\mathit{SR}$ since a sparser pixel $\mathbf{x}$ is obtained.

\subsection{Generalized Orthogonal Matching Pursuit (gOMP)}

The gOMP, presented in Algorithm \ref{alg: GOMP}, enforces a unique $\kappa$-sparse solution for $\mathbf{x}$. The solution is obtained by estimating separately first the indexes of the significant samples in $\mathbf{x}$ and then their respective sample values. The indexes are calculated using the projection $\mathbf{p}$ of the dictionary atoms from $\mathbf{A}$ over the residual from the previous iteration $\mathbf{r}^{i-1}$. The algorithm then takes the $G$ maximum projections in magnitude to determine the $G$ atoms with the highest projections which give at the same time the indexes, in the vector $\mathbf{\Theta}^i$, of the new significant samples to calculate in the current iteration. This step is an important difference with respect to other pursuit greedy algorithms such as the \textit{Orthogonal Matching Pursuit} (OMP) where only one atom is taken at a iteration. The indexes in the vector $\mathbf{\Theta}^i$ are gathered across the iterations in the vector $\mathbf{c}^i$, which collects the significant indexes in $\mathbf{x}$ calculated until the current iteration, and it provides also the indexes of the dictionary atoms with the highest projections allowing to build a subdictionary $\mathbf{B}^i$ which gathers only the atoms with the maximum projections across the iterations. This subdictionary $\mathbf{B}^i$ is used for estimating the significant sample values given in the vector $\mathbf{s}^i$, for the respective indexes collected in $\mathbf{c}^i$. The sparse vector $\mathbf{x}$ is obtained by placing in a $\mathbf{0}$ vector these significant sample values given in $\mathbf{s}^i$ in the respective indexes provided by $\mathbf{c}^i$. However, in order to ensure that the vector $\mathbf{x}$ is $\kappa$-sparse given $\kappa$ as input to the algorithm, only the $\kappa$ maximum values in $\mid\mathbf{x}\mid$ are considered significant and hence their indices determine the atoms indices in the vector $\mathbf{q}^i$ to update the subdictionary $\mathbf{B}^i$ only with the $\kappa$ atoms for the most significant samples. This new subdictionary is used to update the new sample values given in the vector $\mathbf{s}^i$. The final $\kappa$-sparse vector $\mathbf{x}$ is obtained by placing the values in $\mathbf{s}^i$ in the respective indexes given by $\mathbf{q}^i$. The algorithm halts the iterative calculations when the difference $\Delta$ between two consecutive residuals in the $l2$-norm is below a certain threshold $\epsilon$ set experimentally.

\begin{algorithm}[htbp!]
\caption{gOMP}
	\begin{algorithmic}
	\State \textbf{Input:} $\mathbf{y}$, $\mathbf{A}$, $\kappa$,  $G$, $\epsilon$
	\State \textbf{Output:} $\mathbf{x}$ 
	\State \textbf{Initialization:} ,
	$\mathbf{r}^0=\mathbf{y}$, $\mathbf{c}^{0}=\mathbf{\varnothing}$, $\Delta=1$ 
	
    %  while (norm(r) > tolerance && i < min(K, floor(size(A,1)/S)))
		\While{$\Delta \geq \epsilon$}

		\State 1. $\mathbf{p} \gets \mathbf{A}^{*}\mathbf{r}^{i-1}$, \hspace{0.10cm} $\mathbf{\Theta}^{i} \gets$ argmax$_{G}\left(
                    \mathbf{\mid p\mid}\right)$ 
                    
        \hspace{0.15cm}$\mathbf{c}^{i}\gets \mathbf{\Theta}^{i}
        \cup \mathbf{c}^{i-1}$

	\State 2. $\mathbf{B}^i=\mathbf{A}(\forall,\mathbf{c}^{i})$, 
	$\mathbf{s}^i\gets {\mathbf{B}^i}^{+}\mathbf{y}$

	\State 3. $\mathbf{x} \gets \mathbf{0}$, $\mathbf{x}(\mathbf{c}^{i}) \gets {\mathbf{s}}^{i}$
     $\mathbf{q}^{i}$=argmax$_{\kappa}\left(\mid\mathbf{x}\mid\right)$

     \hspace{0.15cm}$\mathbf{B}^i=\mathbf{A}(\forall,\mathbf{q}^{i})$, $\mathbf{s}^i\gets {\mathbf{B}^i}^{+}\mathbf{y}$
     
	\hspace{0.2cm}$\mathbf{x} \gets \mathbf{0}$, $\mathbf{x}(\mathbf{q}^{i}) \gets \mathbf{s}^{i}$
	
	\State 4. $\mathbf{r}^{i}\gets\mathbf{y}-\mathbf{A}\mathbf{x}$, $\quad \Delta \gets \|\mathbf{r}^{i}-\mathbf{r}^{i-1}\|_2$

	\EndWhile

	%This is used to resize parenthesis:
	%$\left(A+B\right)^{-1}$
    
	\end{algorithmic}
	\label{alg: GOMP}
\end{algorithm}

\subsection{Metrics}

\subsubsection{PSNR}
The peak signal-to-noise ratio computes the maximum power of a desired signal with respect to some distorting noise which affects its representation quality~\cite{bib:PSNR_definition}. It is used as a quality measurement to compare the similarity or accuracy of an image $I$ with respect to another reference image $I_{ref}$, and it is calculated as follows:
\begin{equation}
   \mathit{PSNR}\;(I,I_{ref})\;=10\;\cdot\;\log_{10}\left(\frac{R^2}{\mathit{MSE}(I,I_{ref})}\right),
\label{eq:}
\end{equation}

\noindent where $R$ denotes the maximum value in the reference image $I_{ref}$ while $\mathit{MSE}$ is the \textit{Mean Squared Error} between the images $I$ and $I_{ref}$ ~\cite{R_in_PSNR}. The metric can take values in the range of $[-\infty,+\infty]\;\mathrm{dB}$ where the maximum occurs when $I=I_{ref}$. In this work, when comparing the accuracy of one HS image with respect to another reference HS image, the metric is calculated first independently for each spectral band and then the results are averaged.\\

\subsubsection{SSI} The structutural similarity index is based on the human visual system which extracts and processes structural information from the images~\cite{SSIM_orig_paper}. The $\mathit{SSI}$ is used also as a quality measurement metric and it is obtained by comparing the similarities in the lumninance, contrast and structure between small patches in the same locations in the respective images.  The metric can take values in the range of $[0,1]$ where the maximum occurs when $I=I_{ref}$. Analogously to the $\mathit{PSNR}$ metric, the $\mathit{SSI}$ is also calculated bandwise before averaging.

%using small patches in the same locations in the respective images as follows~\cite{SSIM_irena} \textcolor{blue}{(eq. not needed)}:
%\begin{equation}
%     \mathit{SSI}\;(I,I_{ref})\;=l\;(I,I_{ref})\;\cdot\;c\;(I,I_{ref})\;\cdot\;s\;(I,I_{ref}),
% \label{}
% \end{equation}

% \noindent where $l\;(I,I_{ref})$, $c\;(I,I_{ref})$, and $s\;(I,I_{ref})$ denote respectively the similarities in the luminance, contrast and structure between the patches.

%using the mean, standard deviation and cross correlation of the patches in addition to some constants which control the stability together with the robustness of the $\mathit{SSI}$.

\subsubsection{$J$, $t$} The algorithm performance is measured through the number of total convergence iterations, denoted as $J$, to complete the recovery of a full HS image. Additionally, the time needed to recover a complete HS image is given by the recovery time $t$.

\section{Results}
\label{Section: results}

\begin{figure}[!htbp]
\centering
	\begin{subfigure}{0.24\textwidth} % width of left subfigure
	    \includegraphics[width=\textwidth]{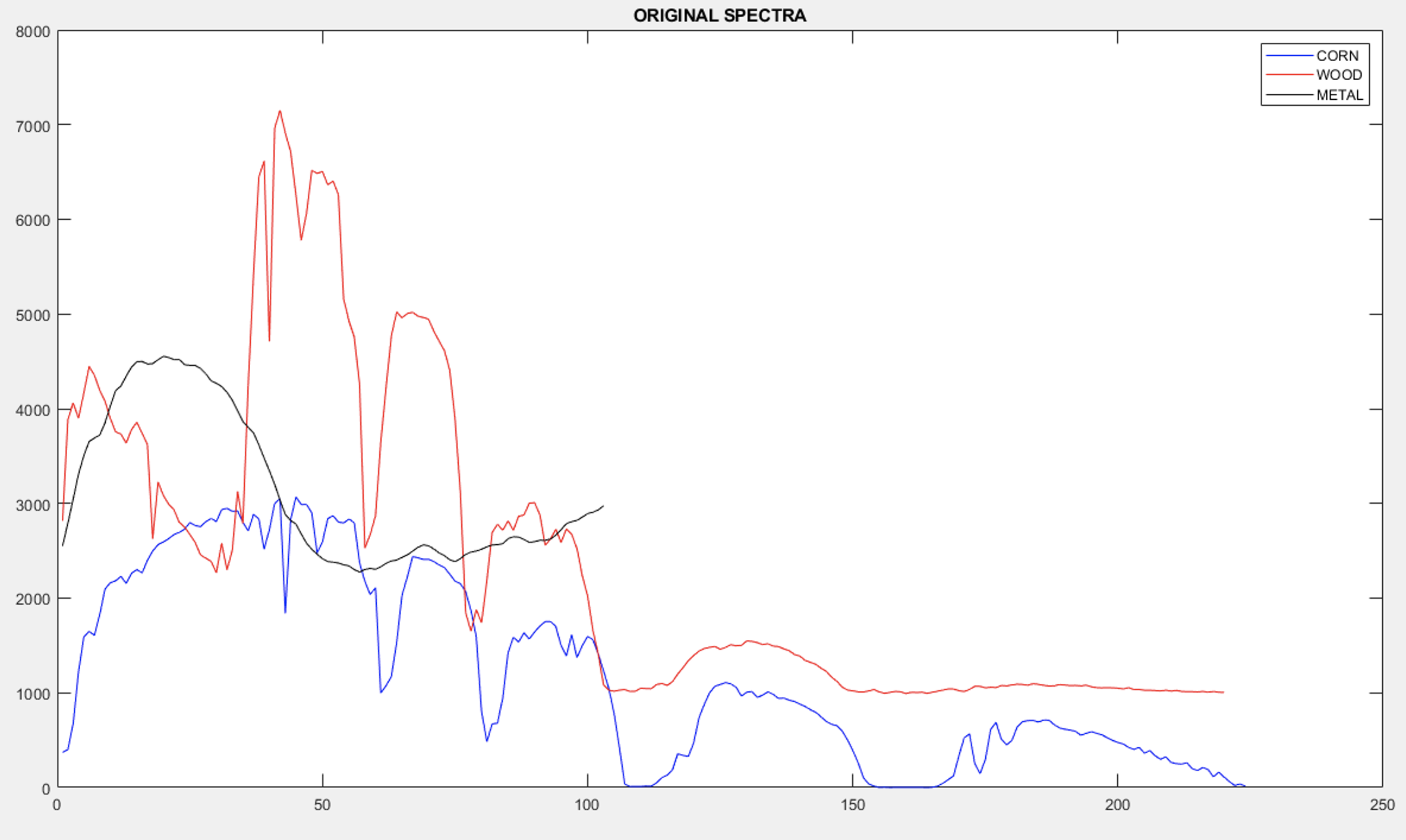}
		\caption{Original Spectra} % subcaption
		\label{SubFig:original}
	\end{subfigure}	
	\begin{subfigure}{0.24\textwidth} % width of left subfigure
	    \includegraphics[width=\textwidth]{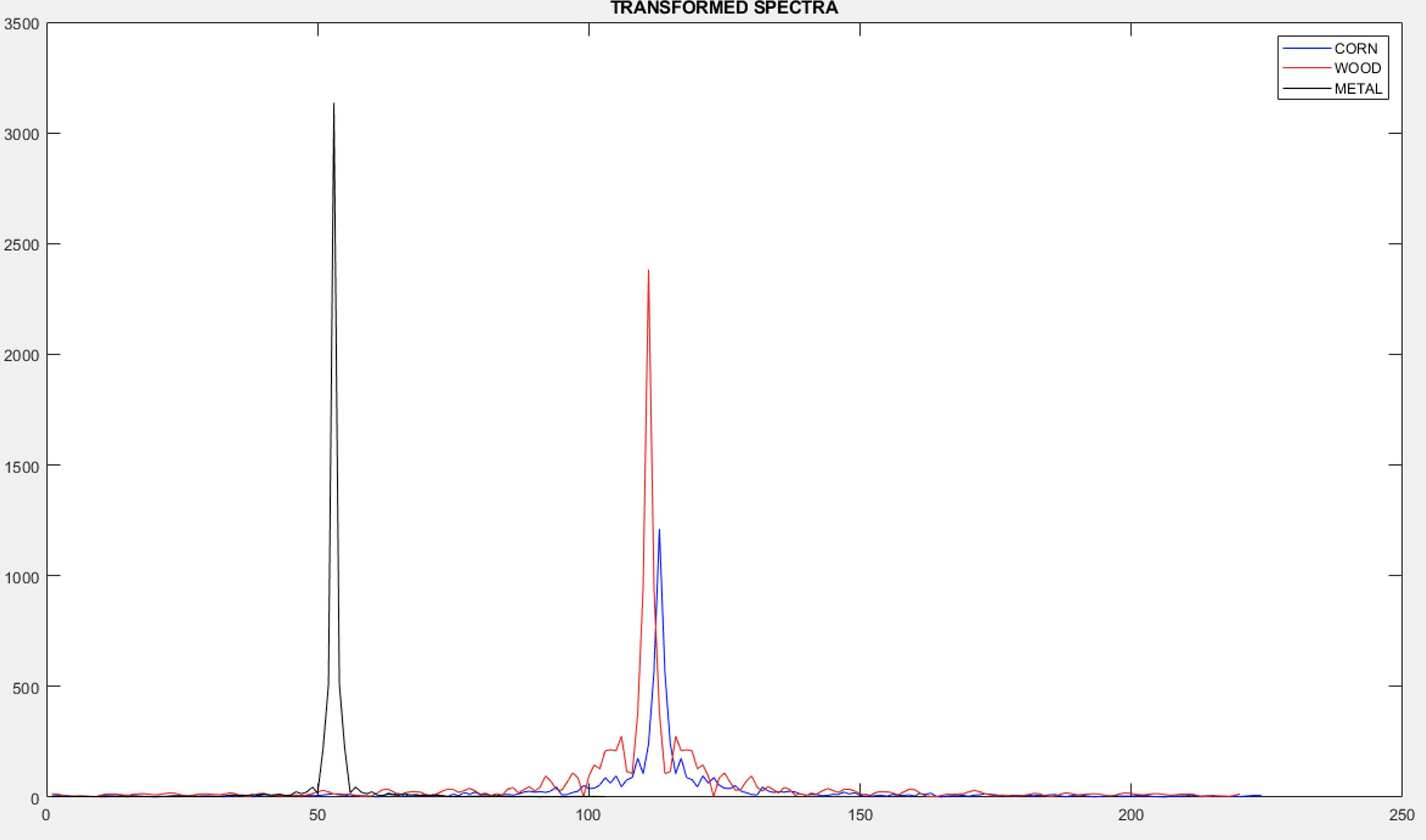}
		\caption{Transformed Spectra} % subcaption
		\label{SubFig:tf_domain}
	\end{subfigure}	
	\begin{subfigure}{0.24\textwidth} % width of left subfigure
	    \includegraphics[width=\textwidth]{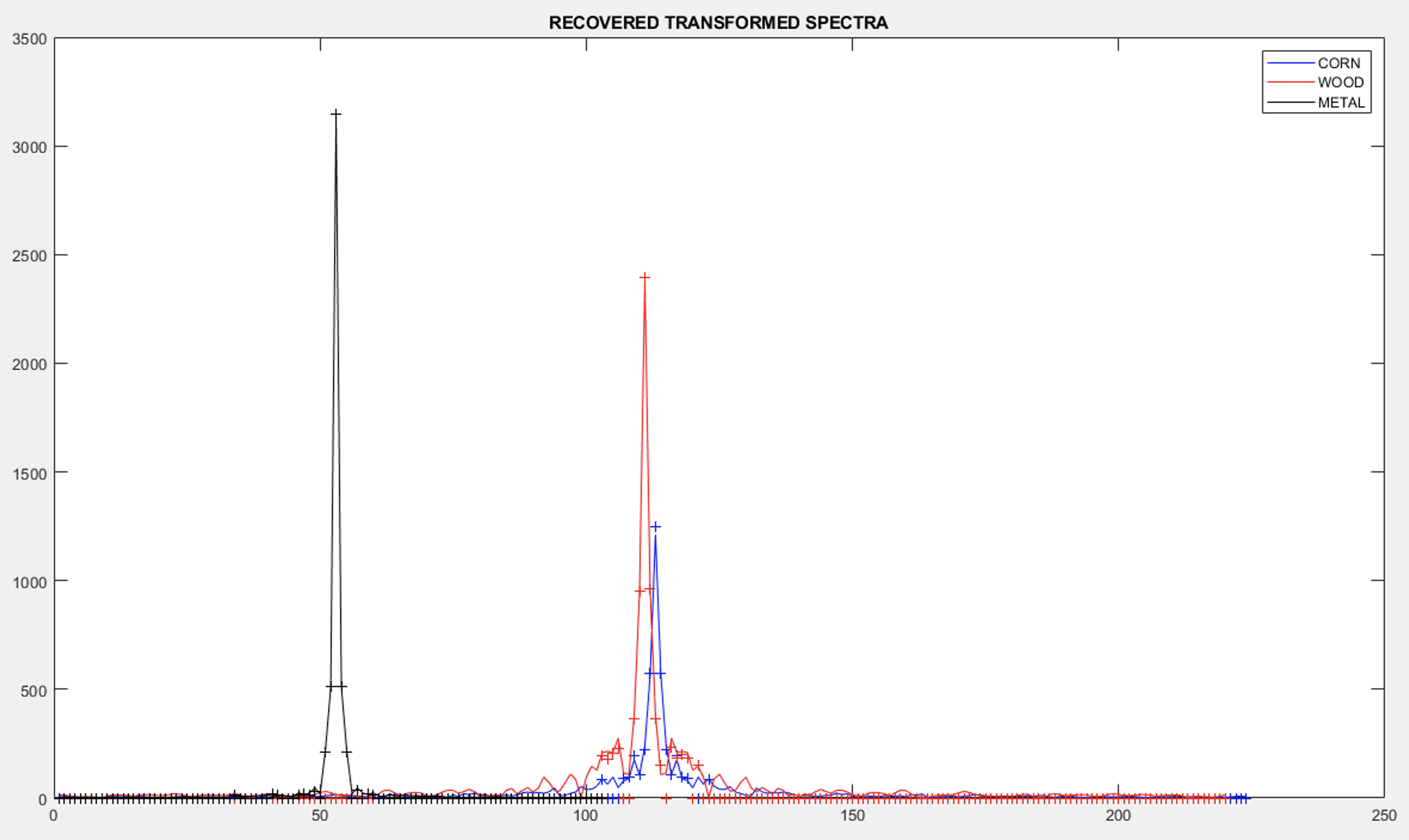}
		\caption{Recovered Transformed Spectra} % subcaption
		\label{SubFig:rec_tf_domain}
	\end{subfigure}		
	\begin{subfigure}{0.24\textwidth} % width of left subfigure
	    \includegraphics[width=\textwidth]{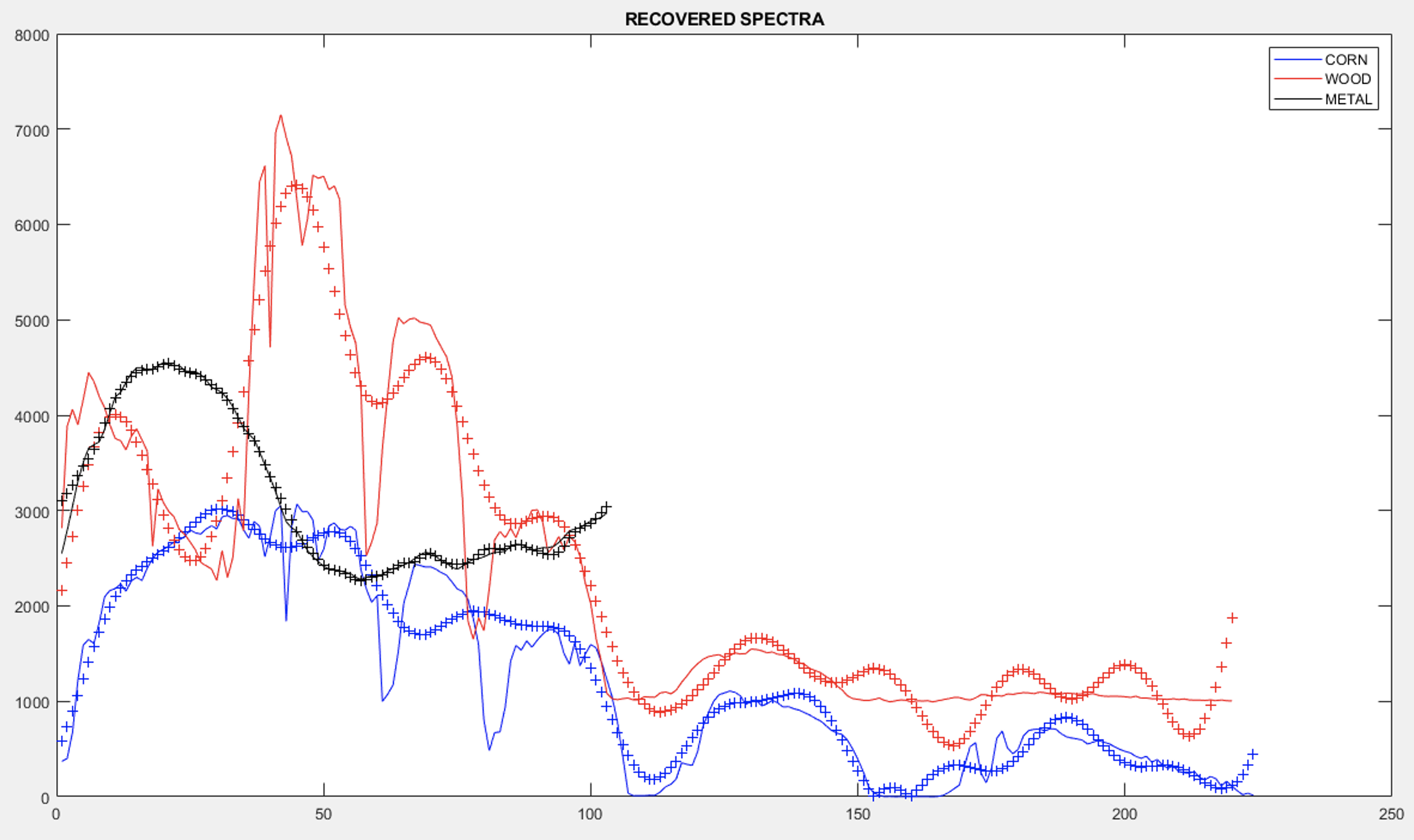}
		\caption{Recovered Spectra} % subcaption
		\label{SubFig:rec_acq_domain}
	\end{subfigure}

		\begin{subfigure}{0.24\textwidth} % width of left subfigure
	    \includegraphics[width=\textwidth]{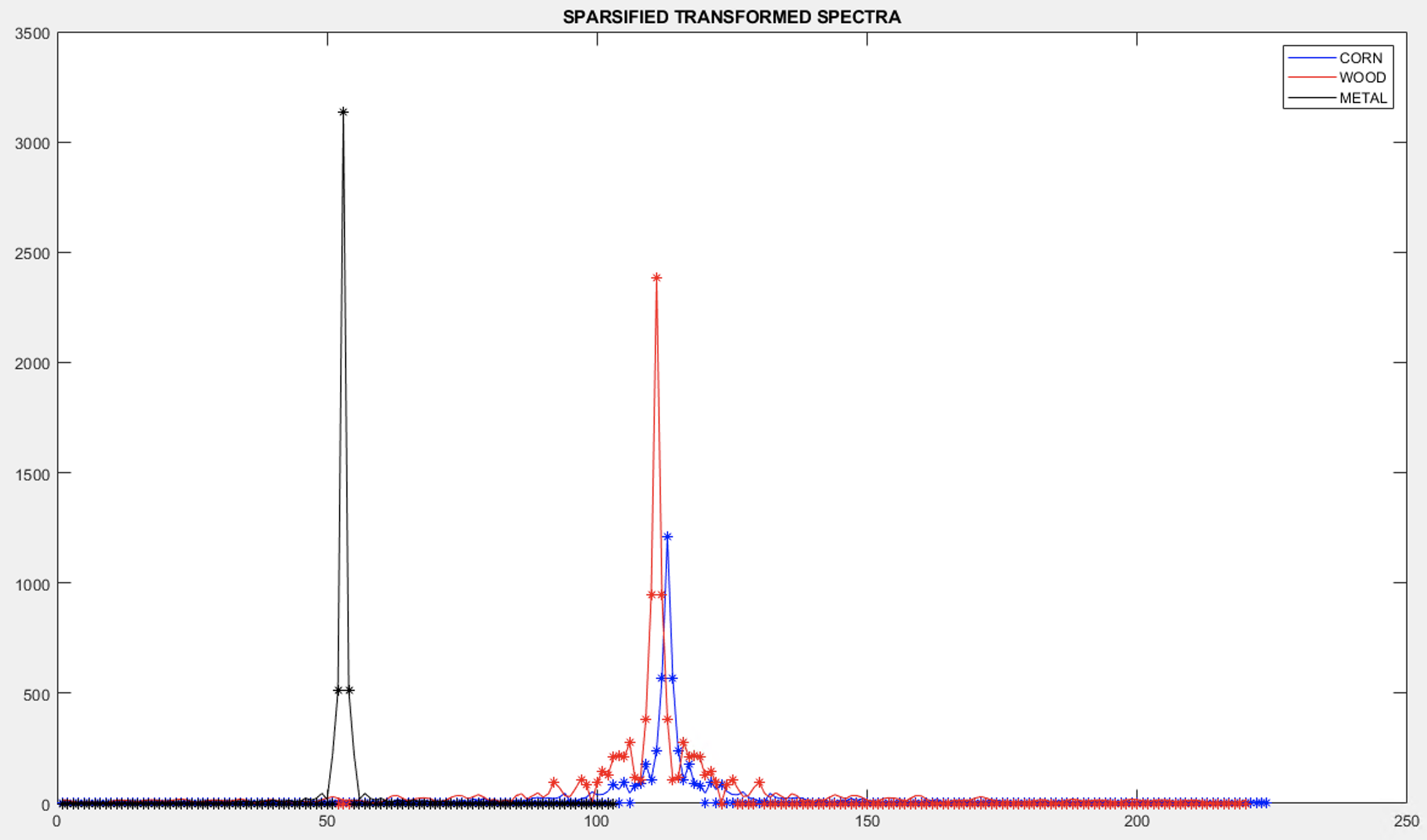}
		\caption{Sparsified Transformed Spectra} % subcaption
		\label{SubFig:spf_tf_domain}
	\end{subfigure}	
	\begin{subfigure}{0.24\textwidth} % width of left subfigure
	    \includegraphics[width=\textwidth]{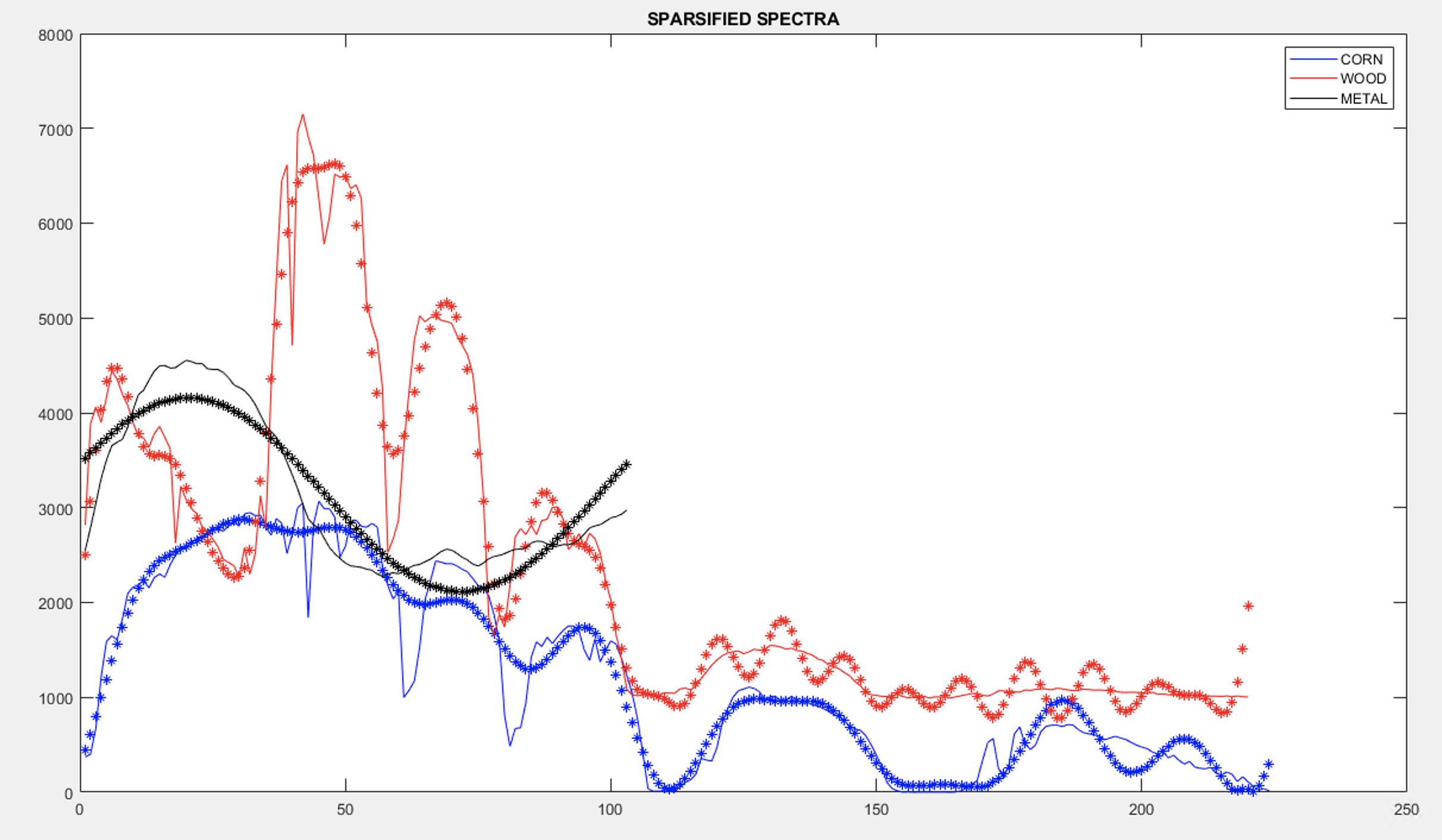}
		\caption{Sparsified Spectra} % subcaption
		\label{SubFig:spf_acq_domain}
	\end{subfigure}	
	\begin{subfigure}{0.24\textwidth} % width of left subfigure
	    \includegraphics[width=\textwidth]{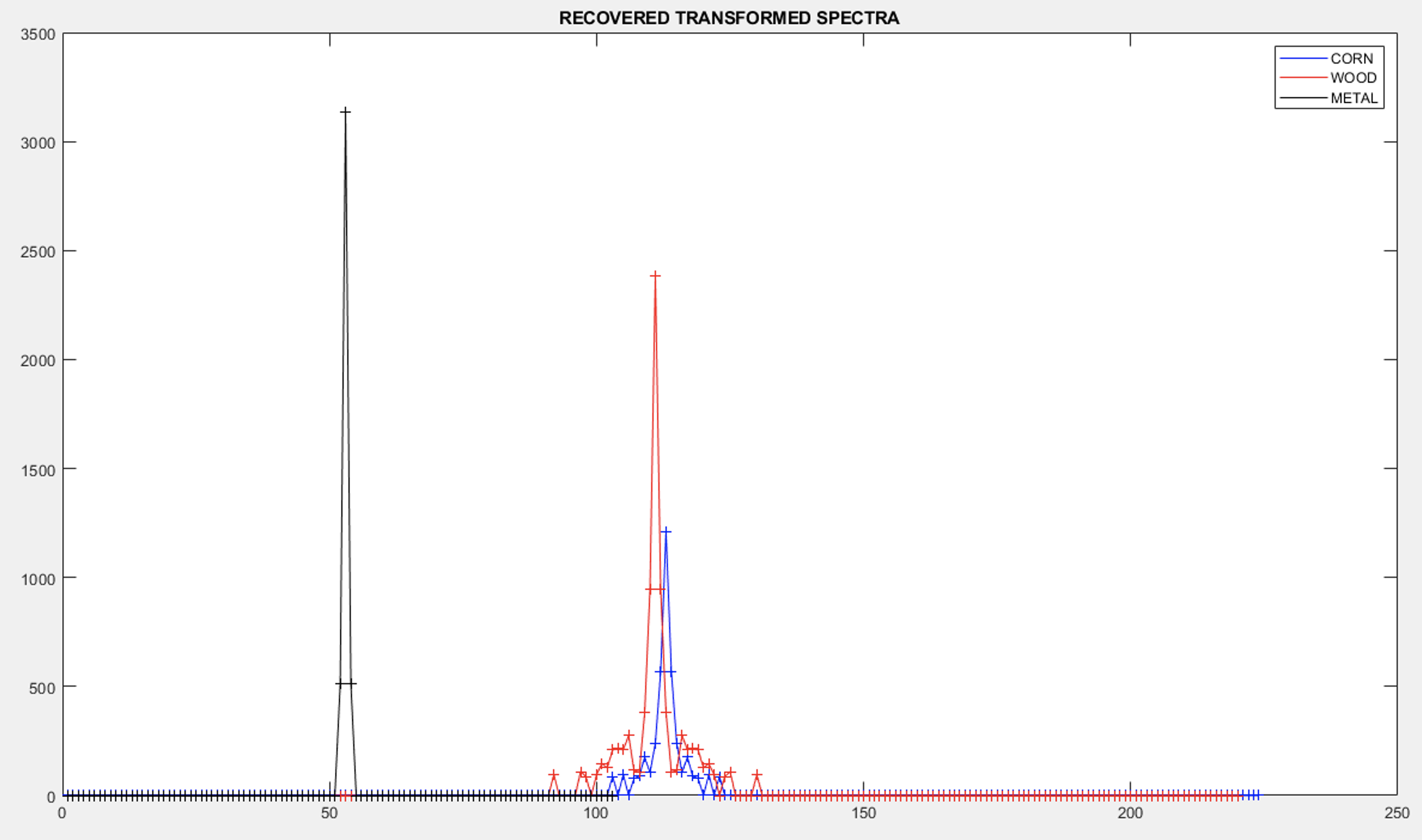}
		\caption{Recovered Transformed Spectra} % subcaption
		\label{SubFig:rec_tf_domain_sparsified}
	\end{subfigure}		
	\begin{subfigure}{0.24\textwidth} % width of left subfigure
	    \includegraphics[width=\textwidth]{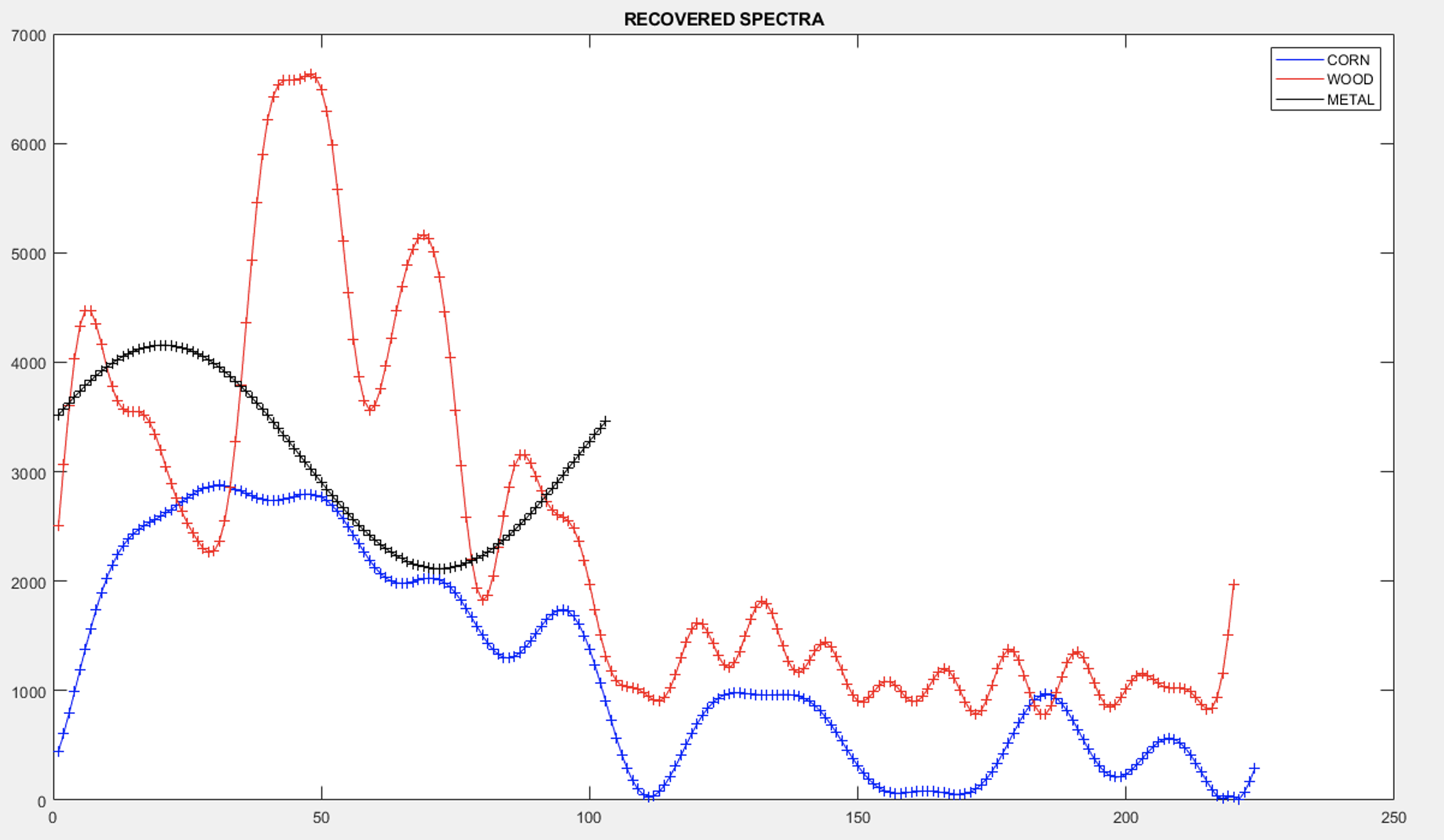}
		\caption{Recovered Spectra} % subcaption
		\label{SubFig:rec_acq_domain_sparsified}
	\end{subfigure}

    \caption{Sparsity Analysis of corn, wood, metal sheet HS pixels}
    \label{Fig:}
\end{figure}
 Experiments are performed in five HS images with dimensions $~X~\times~Y~\times~Z~$ where $X$ and $Y$ are the spatial dimensions and $Z$ is the spectral dimension. The HS data sets used are \textit{Salinas} ($512\times217\times224$), \textit{Pavia University} ($610\times340\times103$), \textit{Indian Pines} ($145\times145\times220$), \textit{Samson} ($95\times95\times156$), and \textit{Cuprite} ($250\times190\times188$). Section~\ref{Section: Sp_analysis} presents a sparsity analysis for targets where their spectra reconstruction is performed by the gOMP algorithm motivating the use of the sparsification stage prior to compression to enhance the spectra recovery. Finally, Section~\ref{Section: HSI_recovery} studies the recovery of the complete HS images in terms of the accuracy and the performance achieved by the gOMP algorithm for different sparsification thresholds.

\subsection{HS Pixels: Sparsity and Sparsification}
\label{Section: Sp_analysis}

\begin{table*}[htbp!]
 \caption{Metrics Analysis for corn, wood, and metal sheet HS pixels}
\label{Table: sparsity_analysis_table}
\begin{tabularx}{\textwidth}{@{}l*{9}{C}c@{}}

\toprule

Target & $K$  & $\mathit{PSNR}(\mathbf{\hat{x}},\mathbf{x})$/ $\mathit{SSI}(\mathbf{\hat{x}},\mathbf{x})$  & $\mathit{PSNR}(\mathbf{\hat{f}},\mathbf{f})$/ $\mathit{SSI}(\mathbf{\hat{f}},\mathbf{f})$ & 

$T$ & 

$\mathit{PSNR}(\mathbf{x_{spf}},\mathbf{x})$/ $\mathit{SSI}(\mathbf{x_{spf}},\mathbf{x})$
&

$\mathit{PSNR}(\mathbf{f_{spf}},\mathbf{f})$/ $\mathit{SSI}(\mathbf{f_{spf}},\mathbf{f})$
&

$\mathit{PSNR}(\mathbf{\hat{x}_{spf}},\mathbf{x_{spf}})$/ $\mathit{SSI}(\mathbf{\hat{x}_{spf}},\mathbf{x_{spf}})$
&

$\mathit{PSNR}(\mathbf{\hat{f}_{spf}},\mathbf{f_{spf}})$/ $\mathit{SSI}(\mathbf{\hat{f}_{spf}},\mathbf{f_{spf}})$ \\ %\addlinespace
\midrule

%\multirow{}{*}{Sparsified}
%\multirow{4}{*}{Sparsified}

Corn  & $15$  & $36.12/0.07$ &  $20.25/0.28$ & $5.5$ & $37.66/0.09$ & $22.07/0.35$ & $332.50/1$ & $315.4/1$\\ %\cline{2-9}

\hline

Wood  & $15$  & $36.67/0.08$ &  $22.95/0.20$ & $3.5$ & $41.35/0.16$ & $27.47/0.34$ & $330.39/1$ & $312.66/1$\\ %\cline{2-9}

\hline

Painted Metal Sheet  & $15$  & $52.62/0.16$ &  $35.18/0.87$ & $7$ & $39.75/0.09$ & $22.86/0.46$ & $332.82/1$ & $314.52/1$\\ %\cline{2-9}

\hline

\bottomrule
\end{tabularx}
\end{table*}

The sparsity of HS pixels is studied using the following three arbitrary targets. Corn from Salinas, wood from Indian Pines, and a painted metal sheet from Pavia University (pixels $348\times52$, $17\times126$, and $169\times130$). Figs.~\ref{SubFig:original} and~\ref{SubFig:tf_domain} present the spectra of the targets in the acquisition and in the transform domain, respectively. The acquisition domain is not sparse since most of the data samples are different from zero. However, it is observed that in the transform domain, for example for the wood, the spectra energy is mostly concentrated within a central peak while the non-significant samples in the sides are close to zero and hence the transform domain constitutes a compressible basis which means that it is not strictly sparse but it can be approximated as sparse. In order to analyse how the gOMP algorithm recovers the spectra in such compressible transform domain, the spectra in the acquisition domain are first randomly subsampled with compression of $2.5\times$ and then subsequently recovered in the transform domain. The algorithm parameters $G$ and $\kappa$ are set experimentally, i.e., $G=2$ as it is found that low $G$ values improve the recovery accuracy achieved by the algorithm and $\kappa=15$ is used as it gives an acceptable $\mathit{PSNR}$ and $\mathit{SSI}$. Figs.~\ref{SubFig:rec_tf_domain} and~\ref{SubFig:rec_acq_domain} show the recovery achieved by the gOMP algorithm in the transform and acquisition domains. It is noticed that the algorithm recovers the painted metal sheet in the acquisition domain with significant higher accuracy compared to the corn and the wood where the algorithm does not capture the rapid signal transitions. This is also shown in Table~\ref{Table: sparsity_analysis_table} which collects the $\mathit{PSNR}$ and $\mathit{SSI}$ metrics measured for the individual targets. The table gives the $\mathit{PSNR}(\mathbf{\hat{x}},\mathbf{x})$ and  $\mathit{SSI}(\mathbf{\hat{x}},\mathbf{x})$ where $\mathbf{\hat{x}}$ and $\mathbf{x}$ denote the recovered and the original spectra in the transform domain, and $\mathbf{\hat{f}}$ and $\mathbf{f}$ are the spectra in the acquisition domain. It is observed that the gOMP algorithm achieves higher $\mathit{PSNR}$ and $\mathit{SSI}$ for the painted metal sheet while comparable accuracy is accomplished for the corn and the wood. Overall, the mentioned lack of sparsity of the targets in the transform domain decreases the accuracy of the spectra recovery since sparse recovery algorithms such as the gOMP algorithm impose strict sparsity of the data to recover in this domain. Table~\ref{Table: sparsity_analysis_table} shows that the accuracy is specially reduced in the acquisition domain where lower $\mathit{PSNR}$ and $\mathit{SSI}$ are obtained compared to the transform domain. In order to ensure that the transform domain constitutes a strictly sparse basis and thus enhance the recovery accuracy achieved by the algorithm, the sparsification stage previously described is applied to the targets spectra prior to compression using the empiric sparsification thresholds $T$ given in the table. Fig.~\ref{SubFig:spf_tf_domain} shows the result of the sparsification in the transform domain where only the samples in the central peak remain due to the significant data they represent, whereas the remaining non-significant samples in the sides of the peak are rounded down to zero. Although the samples in the sides of the peak are considered non-significant as they are relatively small compared to the samples in the central peak, it is observed that when these data samples are approximated to zero, the spectra in the acquisition domain for some targets such as the wood present more signal fluctuations as shown in Fig. ~\ref{SubFig:spf_acq_domain}. Table~\ref{Table: sparsity_analysis_table} shows the recovery accuracy given by the $\mathit{PSNR}$ and the  $\mathit{SSI}$ where $\mathbf{x_{spf}}$ and $\mathbf{f_{spf}}$ are the sparsified spectra in the transform and acquisition domains, and $\mathbf{x}$ and $\mathbf{f}$ are the respective original spectra. It is observed that the values for the $\mathit{PSNR}$ and the $\mathit{SSI}$ in the acquisition domain are not coherently with respect to each other since according to the $\mathit{PSNR}$, the wood is the target with higher metric value compared to the corn and the painted metal sheet which present similar $\mathit{PSNR}$ values, whereas the $\mathit{SSI}$ is higher in the metal sheet with respect to the corn and the wood which have comparable $\mathit{SSI}$ values. In order to study whether the gOMP algorithm can recover the data in the sparsified transform basis, the spectra are compressed $2.5\times$ in the acquisition domain and later recovered with the gOMP algorithm using the parameter $G=2$ and the sparsity level $\kappa$ obtained during the sparsification for each respective threshold $T$ since the in-advance knowledge of the sparsity level of the targets to recover maximizes the recovery accuracy. Figs.~\ref{SubFig:rec_tf_domain_sparsified} and~\ref{SubFig:rec_acq_domain_sparsified} show that the recovery in the sparsified transform and acquisition domains is superior compared to the recovery in the compressible basis in Figs.~\ref{SubFig:rec_tf_domain} and~\ref{SubFig:rec_acq_domain} due to the imposed sparsity in the targets spectra, and shown also in Table~\ref{Table: sparsity_analysis_table} by the $\mathit{PSNR}$ and the $\mathit{SSI}$
where $\mathbf{\hat{x}_{spf}}$ and $\mathbf{x_{spf}}$ denote the recovered and the sparsified spectra in the transform domain, and $\mathbf{\hat{f}_{spf}}$ and $\mathbf{f_{spf}}$ in the acquisition domain. It is observed that close to ideal accuracy is obtained for all the targets. \\

\subsection{Recovery of HS Images}
\label{Section: HSI_recovery}

\begin{table*}[htbp!]
 \caption{Recovery of Compressed HSI Data Cubes : Accuracy and Performance}
\label{Table: summary_table}
\begin{tabularx}{\textwidth}{@{}l*{7}{C}c@{}}

\toprule
Data Set & $T$/$\mathit{SR}$  &
$\mathit{PSNR}/\mathit{SSI}(\mathit{I_{spf}},\mathit{I_{or}})$ &

$\mathit{PSNR}/\mathit{SSI}(\mathit{I_{rec}},\mathit{I_{or}})$ & 

$\mathit{PSNR}/\mathit{SSI}(\mathit{I_{rec}},\mathit{I_{spf}})$ &
 $J$ & $t$\\ %\addlinespace
\midrule

%\multirow{}{*}{Sparsified}
%\multirow{4}{*}{Sparsified}

Salinas  & $5.5$ / $91.22$ &  $25.54$ / $0.62$ & $25.39$ / $0.62$ & $45.53$ / $0.99$ & $1301898$ & $168.72$\\ 

 & $3.5$  /  $87.05$ &  $29.15$ / $0.70$ & $26.62$ / $0.67$ & $32.91$ / $0.94$ & $1813655$ & $337.32$\\   \cline{2-7}

Indian Pines  & $3.5$  /  $88.99$ &  $22.02$ / $0.44$ & $21.94$ / $0.44$ & $46.38$ / $1$ & $302581$ & $42.39$\\ 

  & $2.5$  /  $86.23$ &  $24.94$ / $0.52$ & $23.97$ / $0.50$ & $34.28$ / $0.97$ & $372583$ & $73.13$\\   \cline{2-7} 
 
Pavia Univers. & $7$  /  $94.90$ &  $35.74$ / $0.88$ & $35.60$ / $0.88$ & $54.18$ / $1$ & $1020692$ & $60.89$\\ 

 & $6$  /  $94.02$ &  $36.98$ / $0.88$ & $27.26$ / $0.88$ & $32.50$ / $1$ & $1125133$ & $68.34$\\   \cline{2-7} 
 
Samson & $4.3$  /  $91.52$ &  $30.08$ / $0.77$ & $29.63$ / $0.77$ & $49.57$ / $1$ & $73260$ & $5.78$\\ 

 & $3.8$  /  $90.23$ &  $30.92$ / $0.80$ & $28.75$ / $0.79$ & $37.52$ / $0.98$ & $83111$ & $7.26$\\   \cline{2-7}

Cuprite  & $0.9$  /  $88.73$ &  $40.22$ / $0.93$ & $39.68$ / $0.93$ & $51.64$ / $0.99$ & $627467$ & $87.29$\\ 

  & $0.6$  /  $81.62$ &  $42.45$ / $0.94$ & $38.04$ / $0.87$ & $40.35$ / $0.91$ & $992064$ & $245.68$ \\   \cline{2-7}

\hline

\bottomrule
\end{tabularx}
\end{table*}

\begin{figure}[!htbp]
\centering
	\begin{subfigure}{0.10\textwidth} % width of left subfigure
	    \includegraphics[width=\textwidth]{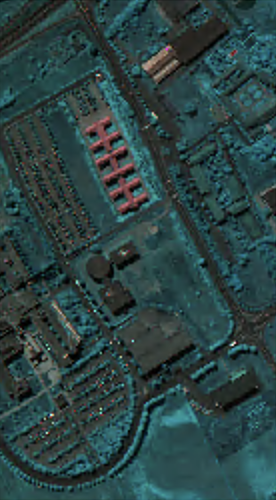}
		\caption{Original} % subcaption
		\label{SubFig:Original}
	\end{subfigure}	
	\begin{subfigure}{0.09\textwidth} % width of left subfigure
	\includegraphics[width=\textwidth]{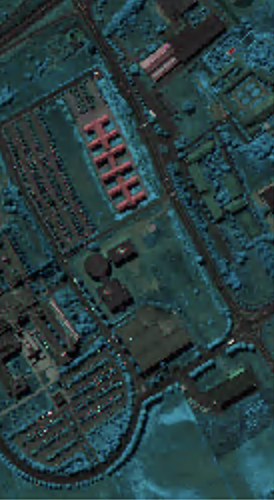}
	\caption{Spf $T=6\;\mathrm{\%}$} % subcaption
		\label{SubFig:PaviaU_Sparsified_T_6_Bands_29_80_97.1}
	\end{subfigure}
	\begin{subfigure}{0.09\textwidth} % width of left subfigure
	\includegraphics[width=\textwidth]{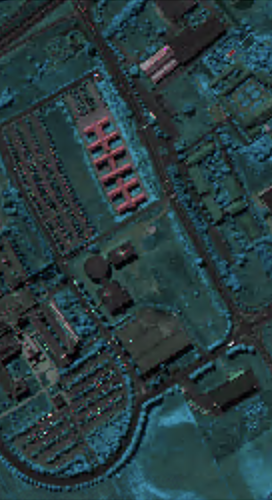}
	\caption{Spf $T=7\;\mathrm{\%}$} % subcaption
		\label{SubFig:PaviaU_Sparsified_T_7_Bands_29_80_97}
	\end{subfigure}
	\begin{subfigure}{0.09\textwidth} % width of left subfigure
	\includegraphics[width=\textwidth]{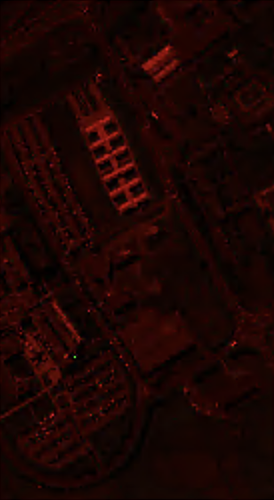}
	\caption{Rec $T=6\;\mathrm{\%}$} % subcaption
		\label{SubFig:PaviaU_Recovered_T_6_Bands_29_80_97}
	\end{subfigure}
	\begin{subfigure}{0.09\textwidth} % width of left subfigure
	\includegraphics[width=\textwidth]{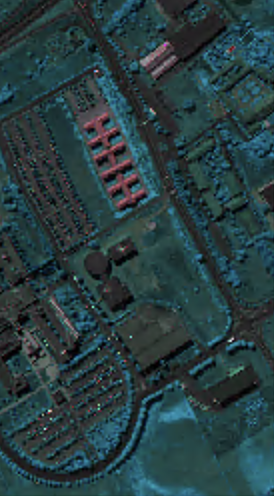}
	\caption{Rec $T=7\;\mathrm{\%}$} % subcaption
		\label{SubFig:PaviaU_Recovered_T_7_Bands_29_80_97}
	\end{subfigure}	
	
    \caption{Pavia University Sparsification and Recovery}
       \label{Fig:general_figure}
\end{figure}

Fig.~\ref{SubFig:Original} shows as an example the false color composite of the bands 29, 80, and 97 from the Pavia University image. Its HS pixels are sparsified with thresholds $T=6\;\%$ and $7\;\%$ before compression, as shown in Figs.~\ref{SubFig:PaviaU_Sparsified_T_6_Bands_29_80_97.1} and~\ref{SubFig:PaviaU_Sparsified_T_7_Bands_29_80_97}. It is subjectively observed that the targets in the bands of the false color composite can be recognised in the respective sparsified images. Table~\ref{Table: summary_table} shows the $\mathit{PSNR}$ and the $\mathit{SSI}$ calculated for all the spectral bands for the sparsified images $\mathit{I_{spf}}$ with respect to the original image $\mathit{I_{or}}$. It is seen that the higher sparsification threshold $T$, which leads to higher $\mathit{SR}$ in the transform domain, gives lower similarity with respect to the original image. Indeed, it is observed that while the Curprite and Pavia University images are the least affected by the sparsification, the Indian Pines image is the most distorted because of the higher noise present in the bands which is considered non-significant data. Fig.~\ref{Fig:general_figure} shows that when the sparsification threshold is raised from $T=6\;\mathrm{\%}$ to $7\;\mathrm{\%}$ ($SR$ is increased to $94.90\;\mathrm{\%}$), then lower metric values are achieved, i.e., as shown in Table~\ref{Table: summary_table} $\mathit{PSNR}/\mathit{SSI}(\mathit{I_{spf}},\mathit{I_{or}})=35.74\;\mathrm{dB}/0.88$ are obtained since the image is sparser for higher thresholds. Figs.~\ref{SubFig:PaviaU_Recovered_T_6_Bands_29_80_97} and~\ref{SubFig:PaviaU_Recovered_T_7_Bands_29_80_97} show the recovery for the respective thresholds achieved by the gOMP algorithm giving $\mathit{PSNR}/\mathit{SSI}\;(\mathit{I_{rec}},\mathit{I_{spf}})\;=32.50\;\mathrm{dB}/1.00$ for $T=6\;\mathrm{\%}$ where $I_{rec}$ is the recovered HS image observed in the figure to be inaccurate. However, the reconstruction is improved when the threshold is raised to $T=7\;\mathrm{\%}$ giving $\mathit{PSNR}/\mathit{SSI}(\mathit{I_{rec}},\mathit{I_{spf}})\;=54.18\;\mathrm{dB}/1.00$, concluding that the algorithm achieves enhanced accuracy when sparser data are recovered. Table~\ref{Table: summary_table} includes additionally the accuracy metrics $\mathit{PSNR}/\mathit{SSI}\;(\mathit{I_{rec}},\mathit{I_{or}})$ which take into account not only the algorithm accuracy but also the sparsification effect. Since the gOMP algorithm assumes that non-significant samples such as noise are zero, the algorithm provides additional noise reduction. Consequently, the results in the table when comparing against the original images instead of the sparsified are not coherent against $T$ since in some data sets the accuracy metrics decrease while increasing in the other ones. Finally, Table~\ref{Table: summary_table} shows the computational requirements by the gOMP algorithm for the different sparsification thresholds. In particular, the table gives the number of iterations demanded to achieve the convergence to recover a complete HS image as well as the total recovery time needed. The table shows that when the sparsification threshold is increased, then the algorithm converges faster and in fewer iterations, concluding that recovering sparser data not only improves the reconstruction accuracy, but it additionally reduces the time needed to reconstruct the HS images.\\

\section{Conclusion}
\label{Section: conclusions}

A sparsity study is conducted for pixels in HS images in the IDFT transform domain. It is concluded that a sparsification stage prior to the image compression is needed to accurately recover the data due to the lack of spectra sparsity in the transform domain. Five HS images compressed $2.5\times$ are recovered with the gOMP algorithm for different sparsification thresholds. The analysis of the recovery accuracy and the performance concludes with the trade-off of higher sparsification thresholds reducing the quality of the original HS images, yet improving the accuracy, reducing the noise and achieving also faster convergence. The suggested further work consists of determining how the presented sparsification stage affects the accuracy of some HS processing techniques such as in semantic segmentation.

\iffalse
\textcolor{blue}{Structure abstract and conclusion:Abstract (1 paragraph): go through the paper and comment the motivation of the problem to solve and why it is solved that way, methodology proposed, brief mention to the most important conclusion or result, remaining conclusion analysis is for the conclusion section itself. Conclusion (1 or 2 paragraph): short summary of what it was done, then summarise main key findings from the results more extensively than in abstract but it's concisely, say implication of this in the field and set further lines of work for it}
\fi

%\clearpage
\bibliographystyle{unsrt} %Sort bibliography according to order in citations
% % \bibliographystyle{plain}
\bibliography{Bibliography}

\section*{Acknowledgment}

The research leading to these results has received funding from the NO Grants 2014 – 2021, under Project ELO-Hyp contract no. 24/2020, and the Research Council of Norway grant no. 223254 (AMOS center of excellence).

\end{document}